\documentclass[10pt,twocolumn,letterpaper]{article}

\usepackage{wacv}
\usepackage{times}
\usepackage{epsfig}
\usepackage{graphicx}
\usepackage{amsmath}
\usepackage{amssymb}
\usepackage{blindtext}
\usepackage{graphicx}
\usepackage{caption}
\usepackage{subcaption}
\usepackage{enumitem}
\usepackage{booktabs}
\usepackage{placeins}
\usepackage{multirow}
\usepackage[pagebackref=true,breaklinks=true,letterpaper=true,colorlinks,bookmarks=false]{hyperref}
\usepackage{cleveref}

\wacvfinalcopy 

\def\wacvPaperID{****} 

\ifwacvfinal\fi
\begin{document}

\title{Multimodal Image Outpainting With Regularized Normalized Diversification}


\author{Lingzhi Zhang\qquad Jiancong Wang\qquad Jianbo Shi \\University of Pennsylvania\\{\tt\small \{zlz, jshi\}@seas.upenn.edu, jiancong.wang@pennmedicine.upenn.edu}}


\twocolumn[{%
\renewcommand\twocolumn[1][]{#1}%
\maketitle
\begin{center}
    \centering
    \includegraphics[width=\textwidth]{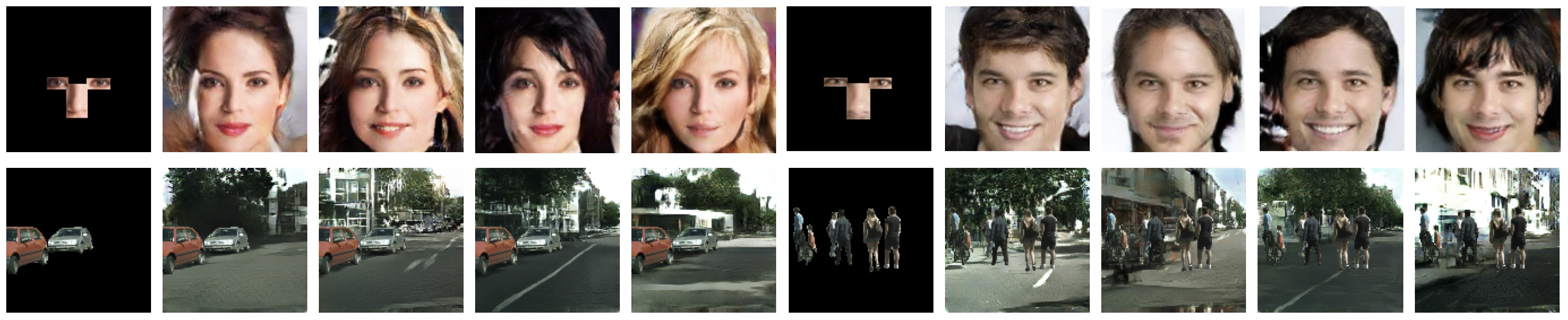}
    \captionof{figure}{Given only a small foreground region, our model can learn to outpaint a set of diverse and plausible missing backgrounds in both face image and street scene image. }
    \label{fig:header}
\end{center}%
}]

\begin{abstract}
In this paper, we study the problem of generating a set of realistic and diverse backgrounds when given only a small foreground region. We refer to this task as image outpainting. The technical challenge of this task is to synthesize not only plausible but also diverse image outputs. Traditional generative adversarial networks suffer from mode collapse. While recent approaches \cite{msgan, ndiv} propose to maximize or preserve the pairwise distance between generated samples with respect to their latent distance, they do not explicitly prevent the diverse samples of different conditional inputs from collapsing. Therefore, we propose a new regularization method to encourage diverse sampling in conditional synthesis. In addition, we propose a feature pyramid discriminator to improve the image quality. Our experimental results show that our model can produce more diverse images without sacrificing visual quality compared to state-of-the-arts approaches in both the CelebA face dataset \cite{celeba} and the Cityscape scene dataset  \cite{cityscape}. Code is available at: \href{https://github.com/owenzlz/Diverse_Outpaint}{https://github.com/owenzlz/DiverseOutpaint}

\end{abstract}

\vspace{-15 pt}
\section{Introduction}


Humans have the ability to hallucinate the possible backgrounds for a given object. For example when one shops for a couch (single foreground object) online, one can imagine how the couch might look inside the living room, one can also imagine how the couch might look in the office (various backgrounds). Is it possible for a machine to do the same? In this paper, we aim to have the machine learn and synthesize a set of diverse and reasonable affordance backgrounds when given a foreground object, especially for cases where large portions of pixels are missing in an image. We refer to this task as image outpainting.

To outpaint the reasonable background for a foreground, the network has to understand the affordance relationship between the foreground and the background. For example, when given features of a person's eyes, the machine needs to infer the possible facial expressions and other facial features of a person. Or given a car or a pedestrian pose, a machine needs to infer the street layout, as shown in Fig.(\ref{fig:header}). While affordance learning \cite{affordance_1, affordance_2, affordance_3, affordance_4, affordance_5, affordance_6, affordance_7} aims to learn how objects interact in an environment, our task focuses on inverse affordance, which hallucinates the environment or background for the objects. Why would this task be useful besides generating interesting images? Some potential applications include facial recognition when large regions of the face are occluded, or synthesizing diverse images for product advertisements. 

\begin{figure*}
\centering
\includegraphics[width=\textwidth]{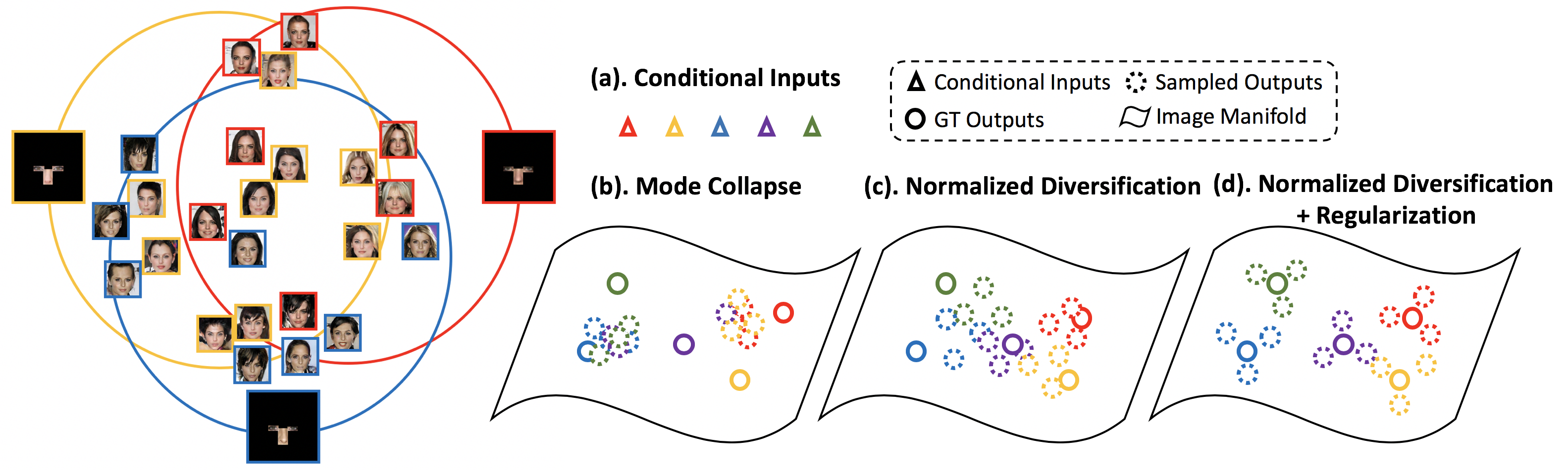}
\caption{\textbf{Motivation} of our diversity regularization. Given conditional inputs (a), generative models could sample many outputs for each conditional input but collapse to a few modes (b). The current solution to this problem (c), normalized diversification \cite{ndiv}, could preserve pairwise distance between sampled outputs for each conditional input, but it does not guarantee that the sampled outputs of different inputs could collapse together. Our solution (d) could not only preserve the pairwise distance of samples for each input but also prevent the samples of different inputs from collapsing together.}

\label{fig:motivation}
\vspace{-10 pt}
\end{figure*}

Intuitively, our task is multimodal common sense learning, which means there exists many possible backgrounds when given only a foreground region. The goal of this paper is to generate not only plausible but also diverse and thorough outputs.
To make this more clear, we draw comparison between out-painting and the more common inpainting, the task of filling in missing pixels in an image. Within inpainting all background and some of the foreground objects are usually given and the semantic relationship between foreground and background objects are pre-determined by the large portion of available pixels. This precludes the generator from hallucinating multiple possible semantic relationship between foreground/background objects. Outpainting, on the contrary, requires to fill in large portion of missing background and there is more degree of freedom in the common-sense semantic relationship that needs to be filled in by the generator. The outpainting therefore imposes higher diversity requirement on the generation framework.


We summarize the contributions of this work as follows. First, we formulated a new image outpainting task and provided a multimodal image synthesis solution. Second, we proposed a new diversity regularization technique to encourage diverse sampling without sacrificing image quality in this conditional synthesis task. In addition, we proposed a novel feature pyramid discriminator to check multi-scale information of outpainted images to improve visual quality. Overall, our proposed method can achieve more diversity and similar or better quality compared to the state-of-the-arts multimodal generative methods in both CelebA \cite{celeba} face dataset and Cityscape \cite{cityscape} street dataset. 

\vspace{-5 pt}
\section{Related Work} 
\vspace{-5 pt}
\subsection{Deep Generative Models}
\vspace{-5 pt}

Deep generative models have produced exciting results. One type of generative models is Generative Adversarial Network (GAN) \cite{gan}. It consists of a generator network (G) and a discriminator network (D). During training, G tries to generate data as similar as the real data while D tries to differentiate the data from the real data. Once the adversarial training reaches an equilibrium, G is able to generate data that is indistinguishable from the real data distribution. Some applications of GAN will be elaborated in section 2.2.

Another popular generative model is variational auto-encoder (VAE), which embeds high-dimensional data into a low-dimensional Gaussian distribution, samples a latent code and decodes it to the output space. The VAE framework is often used in multimodal prediction tasks, where the latent distribution models the uncertainty in the output space. For example, it has been used in multimodal image-to-image translation \cite{bicyclegan, msgan, vaegan}, predicting uncertain future motions \cite{future_motion_1, future_motion_2, future_motion_3}, hallucinating diverse human affordance \cite{human_affordance_1, human_affordance_2} and so on. 

We will discuss the related works that address the mode collapse issue in generative modeling. Mode collapse refers to the degenerate case where the generator produces limited or even single output mode. BourGAN \cite{bourgan} proposes to model the modes as a geometric structure of data distribution in a metric space, and uses mixture of Gaussians to construct latent space in order to map to different modes wihtout collapse in unconditional generation. In conditional generation, mode seeking GAN (MSGAN) \cite{msgan} proposes to maximize the ratio of two sampled images over the their corresponding latent variables as a simple and intuitive diversity regularization. Lastly, normalized diversification \cite{ndiv} proposes to enforce the model to preserve the normalized pairwise distance between the sparse samples from a latent distribution to the corresponding high-dimensional output space. On top of the normalized diversification, we proposed a simple but effective diversity regularization to further encourage diversity in conditional image generation. The experimental results show that our method can generate more diverse images without sacrificing image quality compared to the state-of-the-arts approaches. 


\begin{figure*}
\centering
\includegraphics[width=\textwidth]{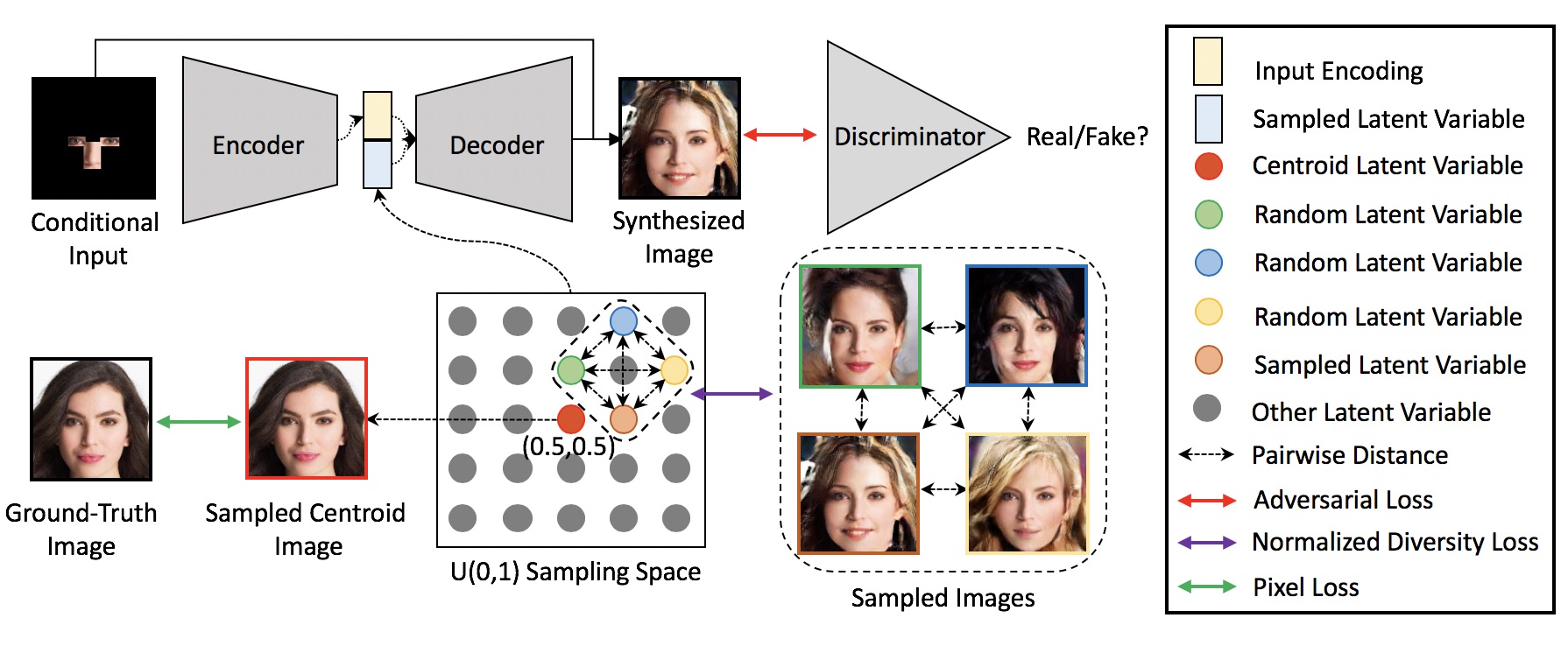}
\caption{\textbf{Model Architecture}. The top part shows the architecture of the network, and the bottom part illustrates how the normalized diversification and diversity regularization are implemented. }
\label{fig:main}
\vspace{-10 pt}
\end{figure*}

\vspace{-5 pt}
\subsection{Conditional Image Synthesis}

Deep generative models have been applied to many conditional image synthesis tasks. In super resolution \cite{SR1, sr2, SR3_style2, sr4, sr5, sr6, sr7}, deep models learn the image texture prior to upsample a low-resolution image into high-resolution version. In style transfer \cite{style1, SR3_style2, style3, style4, style5}, images can be transformed into an arbitrary style while its content being maintained by simultaneously minimizing the content and style loss w.r.t content and style images in feature space. In text-to-image synthesis \cite{text2img_1, text2img_2, text2img_3}, models can synthesize image layout and texture based on the input text. 

Image inpainting, the task of filling parts of missing pixels in an image, is the most similar to our task among conditional image synthesis tasks. Early works \cite{inpaint1, inpaint2} train a deep convolutional network for denoising or inpainting small regions in the image. \cite{inpaint3} proposes to learning useful features using image inpainting with adversarial training. \cite{inpaint4} introduces the global and local discriminators to check the global and local consistency. \cite{inpaint5} iteratively transverses the image manifold to find the closest encoding with respect to the input occluded image and uses it to decode the completed image. More recently, \cite{inpaint6} introduces partial convolution, which is weighted to focus more on the valid regions rather than the hole regions. \cite{inpaint7} first produces a coarse prediction of the missing region in the first stage, and then refines the texture-level details using an attention mechanism by searching for a set of background regions with the highest similarity with the coarse prediction. \cite{inpaint8} proposes to inpaint an image by hallucinating the edge connection in the first stage, and then uses the connected edge map together with occluded image as inputs to produce the final completed image in the second stage. 


Different from the above works, our goal is to produce diverse outputs conditional on a small foreground region. Thus, we only compare our method to the methods that can generate multimodal image solutions.

\section{Methods}
In the image outpainting task, we aim to synthesize a set of plausible and diverse images when given a single foreground input. The previous approaches mostly leverage VAE \cite{vae} to encode a distribution of possible solutions and GAN \cite{gan} to synthesize realistic image. However, these approaches suffer from mode collapse. To build on top of normalized diversification\cite{ndiv}, we proposed a new regularization technique to further encourage image diversity in this conditional image synthesis and a multi-scale discriminator to improve the visual quality. 

\subsection{Normalized Diversification}
In normalized diversification, the generator learns mapping from a uniform latent space to an unknown output space. The key idea is to preserve the normalized pairwise distance of sparse samples between the latent space and the corresponding output space. In details, the Euclidean distance is used as the distance metric, which are shown in Eq.(\ref{1}) and Eq.(\ref{2}). 

\begin{equation}
    d_z(z_i, z_j) = ||z_i - z_j||_2
    \label{1}
\end{equation}

\begin{equation}
    d_{x}(G(z)^i, G(z)^j) = ||G(z)^i - G(z)^j||_2
    \label{2}
\end{equation}

In above, $d_z(z_i, z_j)$ and $d_{x}(G(z)^i, G(z)^j)$ denote the pairwise distance of samples in latent space and output space respectively. We denote $z$ as latent variable, $G$ as generator, $G(z)$ as generated output, and $i,j$ as sample indices. The normalized pairwise distance matrices are further defined as $ D_{ij}^z$, $D_{ij}^{x}$ $\in$ $\mathbb{R}^{N \times N}$ as follows.

\begin{equation}
    D_{ij}^z = \frac{d_z(z_i, z_j)}{\sum_j d_z(z_i, z_j)}
    \label{3}
\end{equation}

\begin{equation}
    D_{ij}^{x} = \frac{d_{x}({G(z)}_i, {G(z)}_j)}{\sum_j d_{x}({G(z)}_i, {G(z)}_j)}
    \label{4}
\end{equation}

During training, we treat the denominator in Eq.(\ref{3}) and Eq.(\ref{4}) as a constant when back-propagating the gradient to the generator network. This ensures that we optimize the absolute pairwise distance, rather than adjusting denominator to satisfy the loss constraint. 

Finally, the normalized diversity constraint is implemented by a hinge loss, where we only penalize the generator when $D_{ij}^{x}$ is smaller than $ D_{ij}^z$ multiplied by a scale factor. 

\begin{equation}
    \mathcal{L}_{ndiv} (x, z) = \frac{1}{N^2-N}\sum_{i=1}^N\sum_{i \ne j}^N max(0, \alpha D_{ij}^z - D_{ij}^{x})
    \label{5}
\end{equation}

As shown in the diversity loss Eq.(\ref{5}), $\alpha$ is the scale hyperparameter, and we do not consider the diagonal elements of the distance matrix, which are all zeros. 

Coupled with the normalized diversity loss, the adversarial loss is used to check the generated diverse outputs are realistic compared to the real data distribution. 
For the discriminator,
\begin{equation}
\begin{split}
    & \mathcal{L}_{D} = \mathbb{E}_{x \sim P_{data}(x)}[ \mbox{min}(0, 1-D(x))] \\& + \mathbb{E}_{z \sim P_{data}(z), z \sim P_z(z)} [\mbox{min}(0, 1+D(G(z)))]
\end{split}    
\end{equation}
For the generator,
\begin{equation}
    \mathcal{L}_{G} = - \mathbb{E}_{z \sim P_z(z)}[(D(G(z)))]
\end{equation}

We refer to $\mathcal{L}_{G}$ as $\mathcal{L}_{adv}$ in the following discussion.

Within our implementation, we use hinge loss to optimize the generator and the discriminator. Spectral normalization\cite{SNGAN} is applied to scale down the weights in discriminator by their largest singular values, which effectively restricts the Lipschitz constant of the discriminator and thus stabilize training. 

\subsection{Diversity Regularization}

In normalized diversification \cite{ndiv}, the diversity loss $L_{div}$ enforces the model to actively explore the output space while the adversarial loss $L_{adv}$ constraints the generated outputs to be reasonable. However, in conditional generation, this framework only enforces the diversity for each conditional input, but do not explicitly prevent sampled outputs of different conditional inputs from collapsing to few modes. In particular, our image outpainting task has a large degree of freedom to synthesize reasonable outputs for a foreground input, and thus the sampled images of a conditional input could be at very diverse locations on the image manifold. Therefore, it is very likely that the sampled images of different conditional inputs could be very visually similar or close on the image manifold. 

To alleviate this issue, we propose a simple yet effective diversity regularization in addition to normalized diversification in this conditional synthesis task. The overall aim is to pull the diverse sampled outputs of different conditional inputs away from each other. Our approach is to impose a hard constraint on the generated outputs decoded from the center point of the uniform latent space, and enforce the generated and corresponding ground-truth outputs to be as similar as possible. This hard constraint is implemented by a pixel-wise Euclidean distance.

\begin{equation}
    \mathcal{L}_{pixel} = ||G(z^*) - x||_2
    \label{9}
\end{equation}

In above, $z^*$ denotes the center point in the uniform latent space and $x$ is the ground-truth image corresponding to the conditional input. A visual demonstration of this insight is shown in Fig.(\ref{fig:motivation}).

\begin{figure}[!h]
\centering    \includegraphics[width=0.47\textwidth]{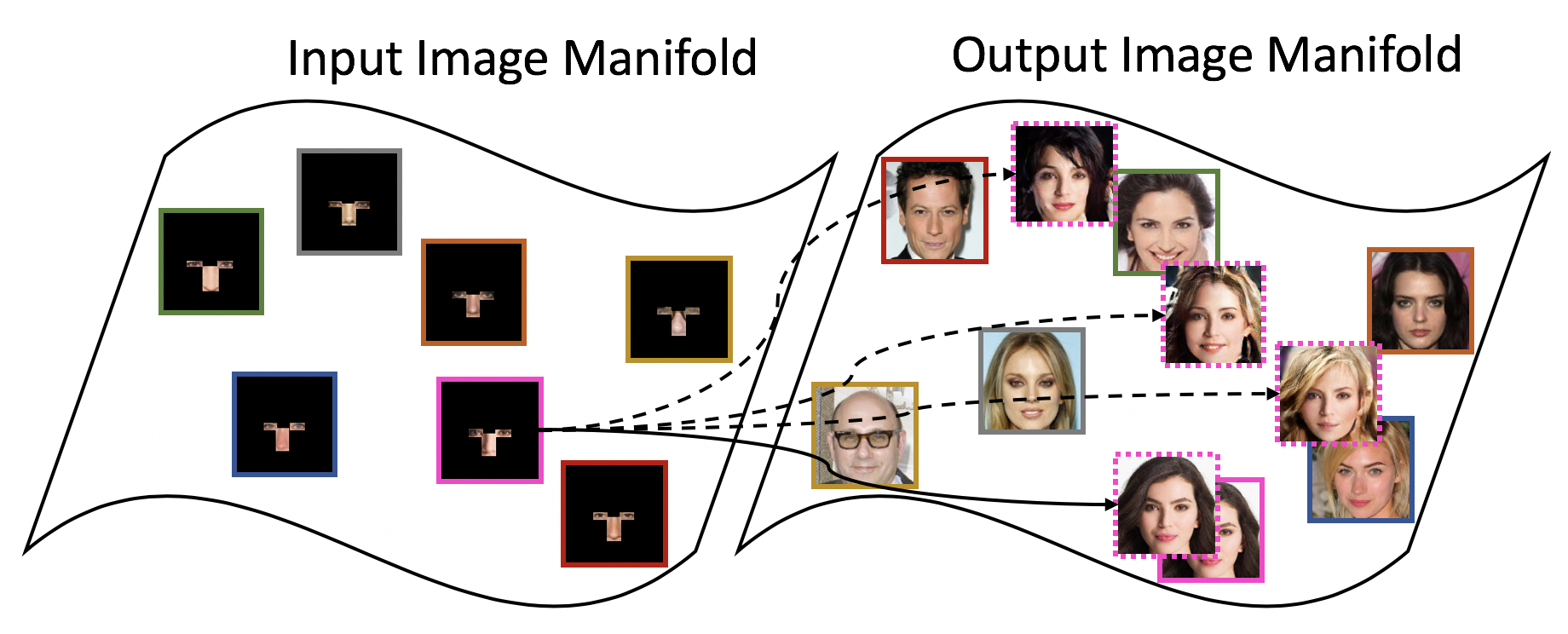}
    \caption{This figure demonstrates the motivation of identity regularization. The solid and dashed arrowed lines indicate the learned mapping with and without this regularization respectively. }
    \label{iden_demo}
\end{figure}

With this diversity regularization, the sampling outputs of each conditional input will be ideally center around its corresponding ground truth output, as shown in Fig.(\ref{iden_demo}). Although this regularization loss might not be fully optimized during training, the sampled outputs of different conditional inputs are pulled away from each other and thus alleviate mode collapse issue we mentioned earlier. The improvement of using this regularization is shown both qualitatively and quantitatively in section 4. 

\subsection{Feature Pyramid Discriminator}

Generative Adversarial Network (GAN) \cite{gan} are commonly used in image synthesis. Prior to GAN, the synthesized images with only pixel reconstruction loss tends to be blurry. The main advantage of GAN is that the discriminator can provide supervisory signal for the generator to synthesize realistic texture of images similar to the real data distribution. Indeed, a recent work \cite{cnn_texture} empirically finds out that CNNs tend to focus on or bias towards visual texture. 

\begin{figure}[!h]
\centering    \includegraphics[width=0.47\textwidth]{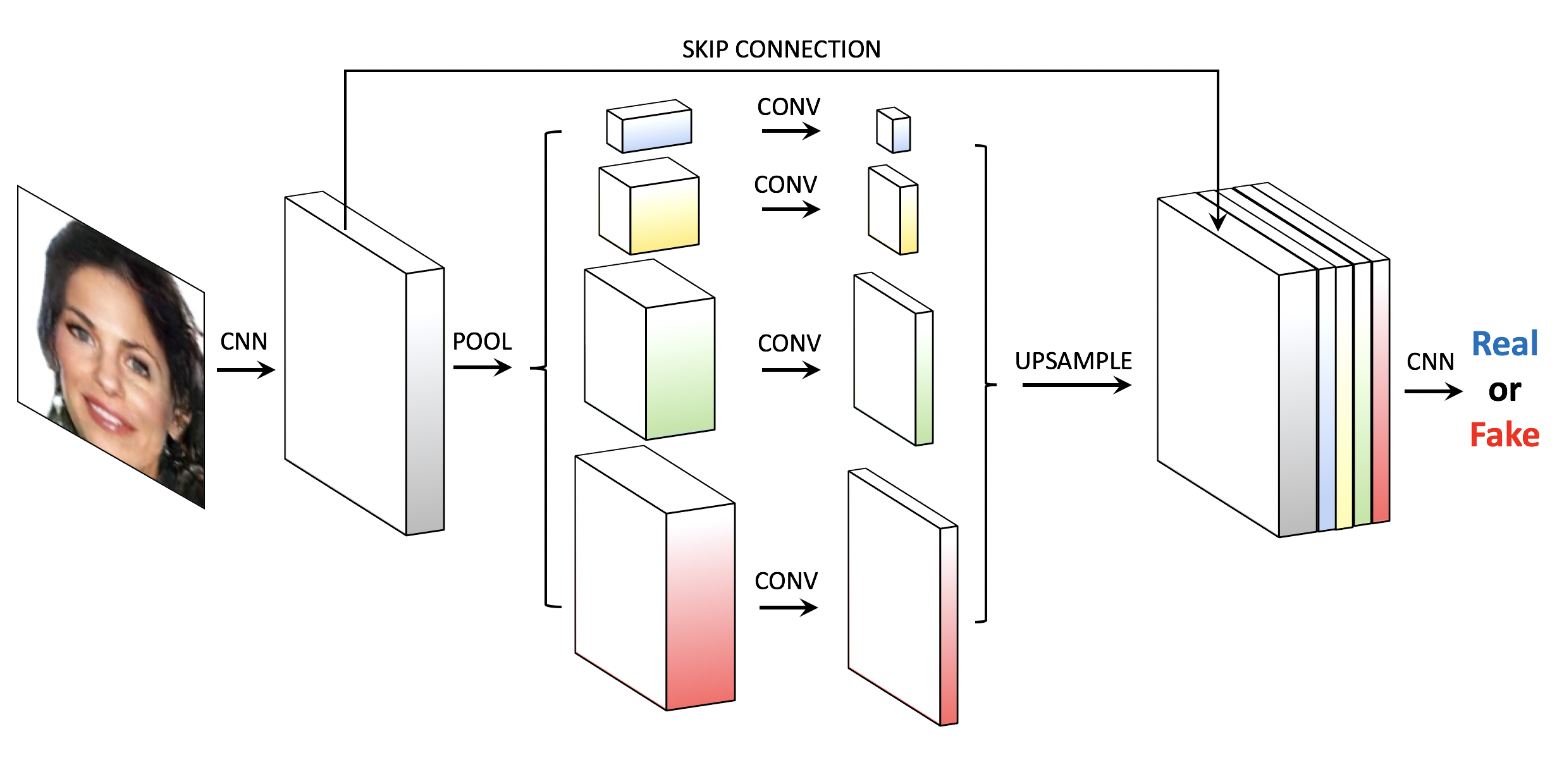}
    \caption{Feature Pyramid Discriminator.}
    \label{multi_disc}
\vspace{-10 pt}
\end{figure}

How to design a discriminator that can check both texture realism and structural plausibility? Our insight is to explicitly design a discriminator network that can focuses on both low-level textures and high-level structural semantics. Inspired by the pyramid scene parsing network PSPNet \cite{pspnet}, we propose to integrate the pyramid pooling module that explicitly computes feature at multiple scales in the discriminator network as above. 

This discriminator first extracts features of an image and downscales the features into multiple scales using average pooling. Then, the downscaled features are squeezed to fewer channel dimensions by a layer of convolution. Finally, the downscaled features are concatenated with original feature and are used jointly to compute the real or fake probability of an image. In summary, our feature pyramid discriminator aims to check multi-scale information of an image and is proven to consistently improve image quality across different datasets, as shown in Table.(\ref{table:quantitative_results}). 

\subsection{Implementation Details}

Our generator network consists of an encoder and a decoder with skip connections at each spatial scale. The discriminator is described in section 3.3. Each convolution and deconvolution layers with stride of 2 are followed by a leaky relu layer with a negative slope of 0.2 and an instance normalization layer. The final output image is combined from the foreground input image and the synthesized output image.

At every step of updates, our model jointly optimizes the diversity loss, the adversarial loss as well as the diversity regularization loss. The overall optimization objective is shown Eq.(\ref{10}). 

\begin{equation}
    \mathcal{L}_{total} = \lambda_1 * \mathcal{L}_{div} + 
    \lambda_2 * \mathcal{L}_{adv} + \lambda_3 * \mathcal{L}_{reg}
    \label{10}
\end{equation}

The hyperparameters $\lambda$ indicate the weights of different optimization objectives. During training, the loss functions are jointly optimized by Adam optimizer with learning rate of 3e-4, beta 1 of 0.5, and beta 2 of 0.99. We use $\lambda_1=0.1$, $\lambda_2=1$, $\lambda_3=5$ for loss weights. 

\section{Experiments}


\subsection{Premilinaries}

\textbf{Datasets.} To generate synthetic training data, we sampled 400 conditional input points from a 2D uniform distribution and computed the corresponding outputs with a discrete non-linear transformation function. In the real image experiments, we used both CelebA \cite{celeba} face dataset and Cityscape \cite{cityscape} street scene dataset. For the face dataset, we center-cropped and scaled the image down to be 128 x 128, and cut out everything but two eyes and nose as inputs. The eyes and nose input region are localized by running a pretrained facial landmark detector Super-FAN \cite{super_fan}. For the street scene dataset, we scaled down the images into 256 x 128 and then cut them by half into 128 x 128 as inputs. We used an instance segmentation network Mask R-CNN \cite{mask_rcnn} to crop out the foreground region as inputs. 

\textbf{Evaluations.} For the synthetic experiments, we evaluated the results by visualizing the sampled output space and calculating the generated data plausibility and diversity quantitatively. For the real image experiments, we did use FID\cite{fid} score to evaluate the image quality and pairwise LPIPS\cite{lpips} score to quantify image diversity. The larger FID score indicates the better image quality, and the larger LPIPS score indicates the better image diversity. Qualitative evaluations are also provided.

\subsection{Baseline Models}

To demonstrate the effectiveness of our method, we compared the results with several state-of-the-arts models as strong baselines.   

\begin{figure*}[!t]
\centering
\includegraphics[width=\textwidth]{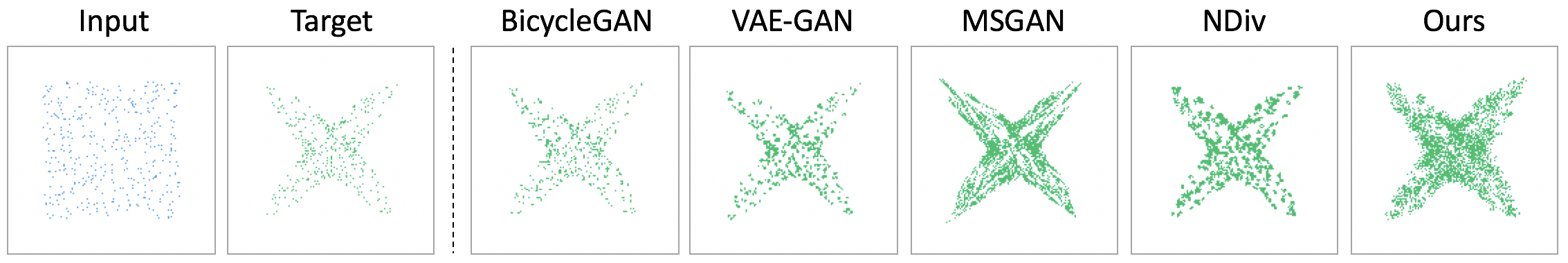}
\caption{Qualitative results on synthetic data.}
\label{synthetic_demo}
\end{figure*}

\begin{table*}[t!]
\centering
\begin{tabular}{llll}
\toprule
 Methods & Frechet Distance $\downarrow$ & Pairwise Distance $\uparrow$ & Number of Modes $\uparrow$ \\
\midrule
 BicycleGAN\cite{bicyclegan} & 1.6477 & 0.0998 & 482\\
 VAE-GAN\cite{vaegan} & \textbf{0.8804} & 0.5898 & 962 \\
 NDiv\cite{ndiv} & 1.2829 & 0.9917 & 1416\\
 MSGAN\cite{msgan} & 1.2985 & 4.0199 & 2041\\
 Ours & 1.0751 & \textbf{4.3332} & \textbf{2481}\\
\bottomrule
\end{tabular}
\caption{Quantitative results on synthetic data.}
\label{table:synthetic_experiment}
\vspace{-10 pt}
\end{table*}

\textbf{cVAE-GAN.} \cite{vaegan} The conditional variational auto-encoder GAN (cVAE-GAN) encodes the input images into a parametric Gaussian distribution and decodes the sampled latent code into the output images, where the model is trained with a reconstruction loss, an adversarial loss and the KL divergence.  

\textbf{BicycleGAN.} \cite{bicyclegan} The BicycleGAN model combines both cVAE-GAN and conditional latent regressor GAN (cLR-GAN) \cite{cLR_GAN}. The key idea is to enforce the connection between latent encoding and output in both directions simultaneously. 

\textbf{MSGAN.} \cite{msgan} Built on top of the conditional GAN \cite{cGAN}, the MSGAN model maximizes the ratio of the distance between generated images with respect to the corresponding latent codes in order to encourage the network to explore more minor modes in the data distribution. 

\textbf{NDiv.} \cite{ndiv} The NDiv model preserves the normalized pairwise distance between the sparse samples from a latent distribution and the generated output space, where the latent space is parameterized using a uniform distribution. 

All of these approaches aim to generate multimodal solutions in conditional image synthesis. Both cVAE-GAN \cite{vaegan} and BicycleGAN \cite{bicyclegan} leverage VAE framework for variational inference but they do not explicitly enforce sampled diverse outputs, and thus mode collapse happens during both training and inference. MSGAN \cite{msgan} proposes to enforces the generated outputs to be as diverse as possible with respect to the corresponding latent codes, but its Gaussian latent distribution puts a strong prior assumption on the output distribution. NDiv \cite{ndiv} does not explicitly prevents the sampled outputs of different conditional inputs from collapsing into a few modes in conditional generation, as we discussed in section 3.2.

\begin{figure*}[t!]
\centering
\includegraphics[width=\textwidth]{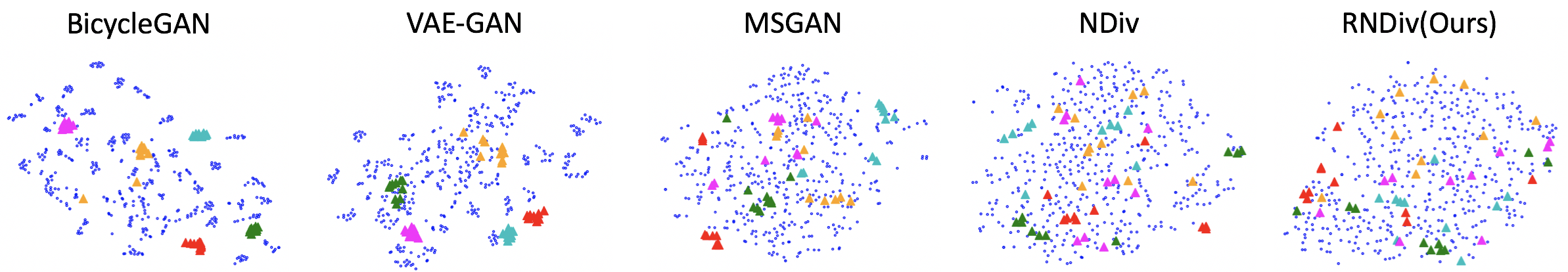}
\caption{To visualize how the sampled images are located on the image manifold, we sampled 10 outpainted images for 100 testing input images in CelebA datast\cite{celeba}. Then, we extracted features for all images using pretrained VGG network\cite{vgg} and ran t-sne\cite{tsne} on the features to visualize the manifold in two dimension. The colored points (pink, orange, green, red, cyan) indicate the sampled images for five specific conditional inputs. Within the same color points, the more spread the points indicate more diverse the sampled outputs for the specific conditional input. The blue dots represent the rest of sampled images. }
\label{fig:tsne}
\end{figure*}

\begin{figure*}[t!]
\centering
\includegraphics[width=\textwidth]{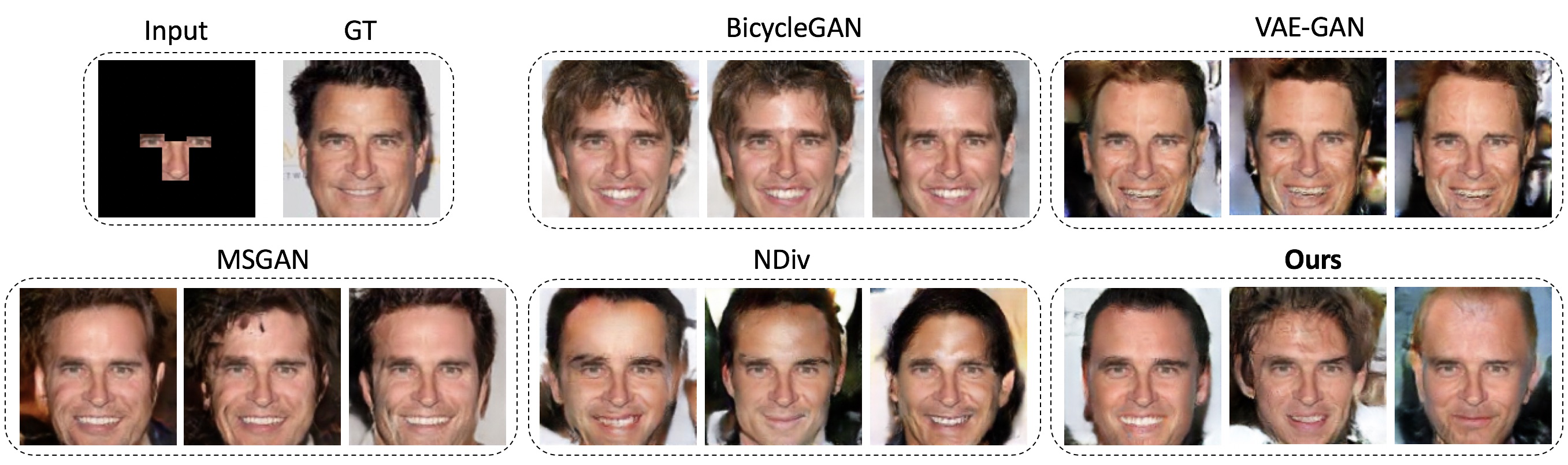}
\caption{For a specific testing input image, we randomly sampled three possible outpainted images with different methods. We intended to demonstrate the diversity levels that each method can produce. Note that the mouths are the same in BicycleGAN\cite{bicyclegan}, VAE-GAN\cite{vaegan}, and MSGAN\cite{msgan}, and hair types are similar in NDiv\cite{ndiv}. In contrast, our method can generate both diverse types of hair styles and mouths. }
\label{fig:visual_comparison}
\end{figure*}


\begin{table*}[t!]
\centering
\begin{tabular}{llllll}
\toprule
 \multirow{2}{*}{Methods} & \multicolumn{2}{c}{CelebA\cite{celeba}} && \multicolumn{2}{c}{CityScape\cite{cityscape}}\\
 \cmidrule{2-3} \cmidrule{5-6}
 & Quality (FID) $\downarrow$ & Diversity (LPIPS) $\uparrow$ && Quality (FID) $\downarrow$ & Diversity (LPIPS) $\uparrow$ \\
\midrule
 BicyleGAN\cite{bicyclegan} & 64.1328 & 0.0927 && 98.8635 & 0.0993 \\
 VAE-GAN\cite{vaegan} & 66.3423 & 0.1754 && 77.8836 & 0.2915  \\
 MSGAN\cite{msgan} & \textbf{56.9978} & 0.2318 && 96.6312 & 0.3096 \\
 NDiv\cite{ndiv} & 68.8545 & 0.3198 && 72.9145 & 0.4238 \\
 Ours (w/o FPD) & 62.0442 & 0.3101 && 66.1893 & 0.4351 \\
 Ours (full model) & 59.4232 & \textbf{0.3274} && \textbf{61.1454} & \textbf{0.4783} \\
\bottomrule
\end{tabular}
\caption{Quantitative results on real images.}
\label{table:quantitative_results}
\end{table*}

\subsection{Synthetic Data Experiment}

To demonstrate the performance of diverse sampling in conditional generation, we start our experiments with a synthetic dataset. This dataset contains a set of sampled points from a uniform space $R^2 \sim U_2(0, 100)$ as inputs and the corresponding output points in a star-shaped space within the same range. We designed a discrete non-linear function to map the input points to the output points, and train the generative models to model such non-linear mapping. Both training and testing contain 400 sampled data points. 

The task is to train a conditional generative model, given an input from the uniform space and a 2-dimension random vector sampled from either a normal distribution (BicycleGAN, VAE-GAN) or from a uniform distribution (NidV, ours), generate a corresponding point in the four-star space. 
In Fig.(\ref{synthetic_demo}), the left two plots indicate the testing conditional inputs and ground truth outputs. The rest of the plots on the right are sampled outputs using different methods. During inference, we sampled ten times for each conditional inputs. Qualitatively, the more distributed the sampled points lie in the output space indicate more diversity the model can produce. As shown in the figure, our model can generate more diverse outputs than other state-of-the-arts methods, since more sampled output points exist. 

\begin{figure*}[t!]
\centering
\includegraphics[width=\textwidth]{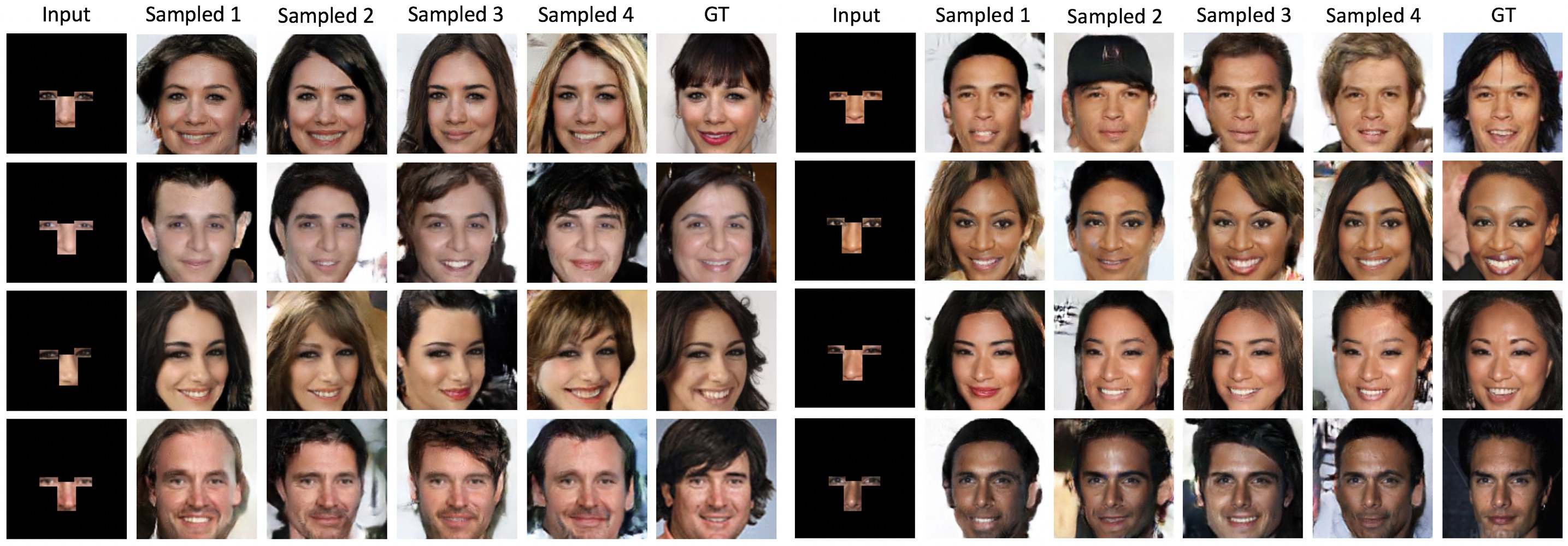}
\caption{Qualitative results of sampled images in CelebA face dataset \cite{celeba}. }
\label{fig:face_results}
\end{figure*}

\begin{figure*}[t!]
\centering
\includegraphics[width=\textwidth]{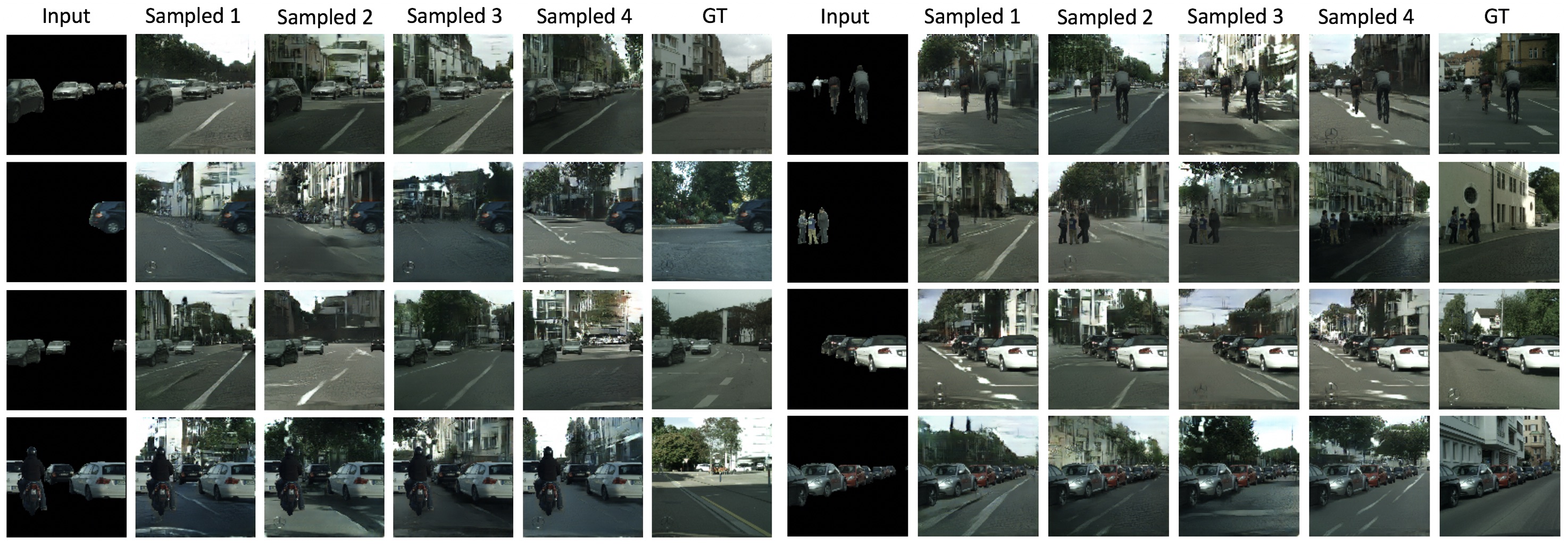}
\caption{Qualitative results of sampled images in Cityscape street scene dataset \cite{cityscape}. }
\label{fig:face_results}
\end{figure*}

In addition, we demonstrate the quantitative comparison study in Table.(\ref{table:synthetic_experiment}). To evaluate the plausibility of the generated outputs, we use the Frechet Distance (FD) to compare the similarity between the generated output distribution and ground truth output distribution. The FD score is computed by averaging across ten batches of sampling outputs with respect to the ground truth. To evaluate the diversity of the generated outputs, we first compute the pairwise distance between the sampled outputs for each conditional input, and then calculate the number of existing output points (modes) in the 2D space. The number of mode is calculated by discretizing the generated output to the closest integer and number of different integers is counted. In conclusion, the quantitative results show that our model can generate most diverse and also plausible outputs compared to the other methods. Although VAE-GAN \cite{vaegan} achieves a slightly better score in terms of plausibility, it only generates limited modes in the output space. 

\subsection{Real Image Experiment}
We conducted image outpainting experiments in both CelebA \cite{celeba} face dataset and Cityscape \cite{cityscape} street scene dataset. In the testing phase, we used 1000 images from both datasets and sample 10 different output images for each conditional input.

To evaluate the generated image diversity, we first show the image manifold of 100 generated output images, as shown in Fig.(\ref{fig:tsne}). The 2D image manifold is obtained by feature extraction using pretrained VGG network \cite{vgg} and t-SNE \cite{tsne} dimensionality reduction. In Fig.(\ref{fig:tsne}), we use five different color (pink, orange, green, red, cyan) to indicate sampled output images for five specific conditional input. This is intended to show the locations of diverse generated output images on the image manifold for the same conditional input. BicycleGAN \cite{bicyclegan} and VAE-GAN \cite{vaegan} both have obvious mode collapse issue, since the sampled images are mostly clustered together into few modes, which leads to the big "holes" in the image manifold. With explicitly diversity losses, both MSGAN \cite{msgan} and NDiv \cite{ndiv} can generate much more diverse outputs, but some generated images still collapse together, such as the red points. In contrast, our method can generate the most diverse outputs compared to these methods. On the image manifold, almost all the randomly sampled outputs stays away from each other and thus results in a more expanded manifold than the others. A set of generated images from different methods are shown in Fig.(\ref{fig:visual_comparison}). 


In the quantitative evaluation, we use the Frechet Inception Distance (FID) \cite{fid} to measure the image quality and the pairwise Learned Perceptual Image Patch Similarity (LPIPS) \cite{lpips} to measure the image diversity. The pairwise LPIPS score is computed between ten sampled genereated images and is averaged across the entire testing set. As shown in Table.(\ref{table:quantitative_results}), our model can generate substantially more diverse images and achieve similar or better image quality compared to the state-of-the-arts methods on both datasets. Note that our proposed feature pyramid discriminator (FPD) improves the image quality and our proposed diversity regularization improves the image diversity consistently in both datasets. 


\section{Conclusion}

In this paper, we formulated the image outpainting task, which aims to synthesize a set of realistic and diverse backgrounds when given only a small existing region. Based on the normalized diversification, we proposed a new regularization technique to further resolve mode collapse issue in this conditional image synthesis task. We also proposed a feature pyramid discriminator to improve the visual quality of generated images by checking image information at multi-scale. Finally, we demonstrated the effectiveness of our method compared to the state-of-the-arts methods in terms of both image diversity and quality in both synthetic dataset and two real-world datasets. In the future, we believe that our work could be applied to occluded face recognition for forensic purpose or common image editing, and could potentially extended to understand object affordance within detection/segmentation tasks. 

{\small
\bibliography{egbib}
}

\end{document}